\documentclass[conference]{IEEEtran}
\IEEEoverridecommandlockouts
\usepackage{cite}
\usepackage{amsmath,amssymb,amsfonts}
\usepackage{algorithmic}
\usepackage{graphicx}
\usepackage{textcomp}
\usepackage{xcolor}
\usepackage{makeidx}
\usepackage{booktabs}
\usepackage{bbding}
\usepackage{multirow}
\usepackage{url}
\usepackage{floatrow}
\usepackage[misc]{ifsym}
\usepackage{subfigure}
\newfloatcommand{capbtabbox}{table}[][\FBwidth]
\def\BibTeX{{\rm B\kern-.05em{\sc i\kern-.025em b}\kern-.08em
    T\kern-.1667em\lower.7ex\hbox{E}\kern-.125emX}}
\begin{document}

\title{Flow-Mixup: Classifying Multi-labeled Medical Images with Corrupted Labels}
\author{
\makebox[.5\linewidth]{Jintai Chen}\\
\IEEEauthorblockA{
\textit{College of Computer Science and Technology}\\
\textit{Zhejiang University} \\
Hangzhou, China \\
jtchen721@gmail.com}\\
\and \makebox[.5\linewidth]{Hongyun Yu}\\
\IEEEauthorblockA{
\textit{College of Computer Science and Technology}\\
\textit{Zhejiang University}\\
Hangzhou, China\\
yuhongyun777@zju.edu.cn}\\
\and \makebox[.5\linewidth]{Ruiwei Feng}\\
\IEEEauthorblockA{
\textit{College of Computer Science and Technology}\\
\textit{Zhejiang University}\\
Hangzhou, China \\
ruiwei\_feng@zju.edu.cn}\\
\and \makebox[.5\linewidth]{Danny Z. Chen}\\
\IEEEauthorblockA{
\textit{Department of Computer Science and Engineering} \\
\textit{University of Notre Dame}\\
Notre Dame, IN 46556, USA \\
dchen@nd.edu}\\
\and \makebox[.5\linewidth]{Jian Wu~\textsuperscript{\Letter}}\\
\IEEEauthorblockA{
\textit{Zhejiang University School of Medicine} \\
\textit{Zhejiang University}\\
Hangzhou, China \\
wujian2000@zju.edu.cn}
}
\maketitle

\begin{abstract}
In clinical practice, medical image interpretation often involves multi-labeled classification, since the affected parts of a patient tend to present multiple symptoms or comorbidities. Recently, deep learning based frameworks have attained expert-level performance on medical image interpretation, which can be attributed partially to large amounts of accurate annotations. However, manually annotating massive amounts of medical images is impractical, while automatic annotation is fast but imprecise (possibly introducing corrupted labels). In this work, we propose a new regularization approach, called Flow-Mixup, for multi-labeled medical image classification with corrupted labels. Flow-Mixup guides the models to capture robust features for each abnormality, thus helping handle corrupted labels effectively and making it possible to apply automatic annotation. Specifically, Flow-Mixup decouples the extracted  features by adding constraints to the hidden states of the models. Also, Flow-Mixup is more stable and effective comparing to other known regularization methods, as shown by theoretical and empirical analyses. Experiments on two electrocardiogram datasets and a chest X-ray dataset containing corrupted labels verify that Flow-Mixup is effective and insensitive to corrupted labels.
\end{abstract}


\section{Introduction}\label{introduction}
Medical image classification is critical in clinical practice (e.g., for early detection of diseases). However, medical image classification is still a challenging task due to large intra-class variations, blurred boundaries between abnormalities, inconclusive abnormality patterns, etc. Furthermore, it is common for a patient to suffer multiple symptoms simultaneously, which present several kinds of abnormalities and some complex comorbidities in the medical images. Therefore, medical image interpretation is commonly a multi-labeled image classification process. Recently, supervised deep learning models have attained high performance thanks to large amounts of well-annotated data for model training. However, it is time-consuming to annotate medical images manually by medical experts, while automatic annotation (e.g., automatically extracting labels from reports~\cite{Wang2017CVPR}) is fast but possibly introduces considerable corrupted (incorrect) labels.\\
\indent Many methods, such as model ensemble~\cite{CheXNeXt}, weighted loss function~\cite{DNetLoc}, and label hierarchy~\cite{chen2019deep}, were widely utilized in multi-labeled medical image classification. However, dealing with corrupted labels of multi-labeled medical images was rarely studied, which is a basic issue for using automatically annotated labels. It was proved that regularization methods could hinder the memorization of models with the generalization ability preserved~\cite{arpit2017closer}, which is advantageous to tackling label corruption. Many known regularization methods (e.g., Mixup~\cite{MIXUP}, Manifold Mixup~\cite{manifoldmixup}) were proposed for single-output tasks, but could not meet the needs of multi-output tasks~\cite{xu2019survey}. Besides, the existence of complex correlations among abnormalities was confirmed in literature~\cite{graphx,Wang2017CVPR}, which required additional considerations in model training. Thus, multi-labeled medical image classification with corrupted labels is a challenging problem and requires further research effort.\\
\indent To this end, in this paper, we propose a new regularization approach called Flow-Mixup for multi-labeled medical image classification with label corruption. Specifically, we introduce a new dimension called ``flow dimension'' for the feature tensors in hidden states and apply a novel Mixing module to a selected hidden state\footnote{In this paper, we denote a ``hidden state'' as the general output of a model layer, and a ``feature'' refers to a particular representation of some data.}. Thus, the model layers ahead of the selected hidden state are restricted to learning a nonlinear function while the subsequent layers are restricted to learning a linear function. Flow-Mixup guides the nonlinear part to decouple the complex features (where the features of abnormalities are correlative) into abnormality-specific features before feeding the features to the linear part. The decoupling is guaranteed as the linear function requires its input features to lie in a linearly separable space. 
We compare Flow-Mixup with Mixup~\cite{MIXUP} and Manifold Mixup~\cite{manifoldmixup} to highlight the characteristics of Flow-Mixup.\\
\indent This work makes three main contributions:
\begin{enumerate}
    \item We propose a new regularization method called Flow-Mixup for multi-labeled medical image classification, and show that Flow-Mixup is insensitive to corrupted labels.
    \item We compare Flow-Mixup with Mixup~\cite{MIXUP} and Manifold Mixup~\cite{manifoldmixup}, and show that the ``correlation conflicts'' phenomenon and the ``distribution shift'' phenomenon occur with using Mixup or Manifold Mixup.
    \item Experiments on several multi-labeled medical image classification datasets with corrupted labels verify that our Flow-Mixup outperforms known regularization methods.
\end{enumerate}
\section{Related Work}\label{relatedworks}
\subsection{Multi-labeled Medical Image Classification}
\indent Various automatic medical image interpretation applications involve multi-labeled image classification tasks, such as chest X-ray (CXR) interpretation~\cite{cicero2017training,Wang2017CVPR,graphx,CheXNeXt,subspace}, electrocardiogram (ECG) monitoring~\cite{shen2019ambulatory,kachuee2018ecg,golany2019pgans}, comorbidity identification of Alcohol Use Disorder and human immunodeficiency virus infection~\cite{adeli2018multi}, bone fracture type diagnosis~\cite{lee2020long}, etc. To better handle multi-labeled classification tasks, a new loss function was proposed to guide deep learning models to search the subspace of abnormality features~\cite{subspace}, and label hierarchy~\cite{chen2019deep} and matrix completion~\cite{adeli2018multi} methods were also used in correlative feature refinement. In~\cite{uncertainty}, an approach was especially designed to calculate the uncertainty of automated diagnosis. Abnormality location perception was considered in~\cite{DNetLoc,infomask} for CXR image classification. Also, adversarial learning approaches were designed for data augmentation and disease severity assessment in CXR~\cite{xing2019adversarial,lanfredi2019adversarial} and ECG~\cite{golany2019pgans} classification. A large dataset~\cite{Wang2017CVPR} catalyzed multi-labeled classification methods on CXR images. However, the labels in this dataset were mined from radiology reports by natural language processing (NLP) and the text-mined labels were somewhat corrupted. Most of the known supervised multi-labeled classification methods focused on tackling feature correlations among abnormalities but few of them considered label corruption.
\subsection{Regularization Methods}
\indent Regularization methods are useful for dealing with label corruption~\cite{arpit2017closer}. Kurmann et al.~\cite{subspace} managed to drive class-specific features into different affine subspaces and enlarge the distances between the subspaces. This method outperformed the vanilla methods in multi-labeled CXR image classification. Many data augmentation methods were used to deal with multi-labeled medical image classification~\cite{graphx,DNetLoc,CheXNeXt}, which had similar effect as regularization methods. The state-of-the-art regularization methods for single-labeled classification are Mixup~\cite{MIXUP,analysisMixup} and Manifold Mixup~\cite{manifoldmixup}, by introducing linear constraints into the models. However, both of them are not very suitable to multi-labeled classification because of the ``correlation conflicts'' and ``distribution shift'' phenomena (discussed in Sec.~\ref{sec:analysis}). Mixup ignored the feature correlations among abnormalities while Manifold Mixup was often unstable in training. In this paper, we propose Flow-Mixup for multi-labeled medical image classification, which avoids the drawbacks of Mixup and Manifold Mixup.
\section{Approach}\label{sec:approach}
\begin{figure*}[tb]
    \centering
    \includegraphics[width=0.9\textwidth]{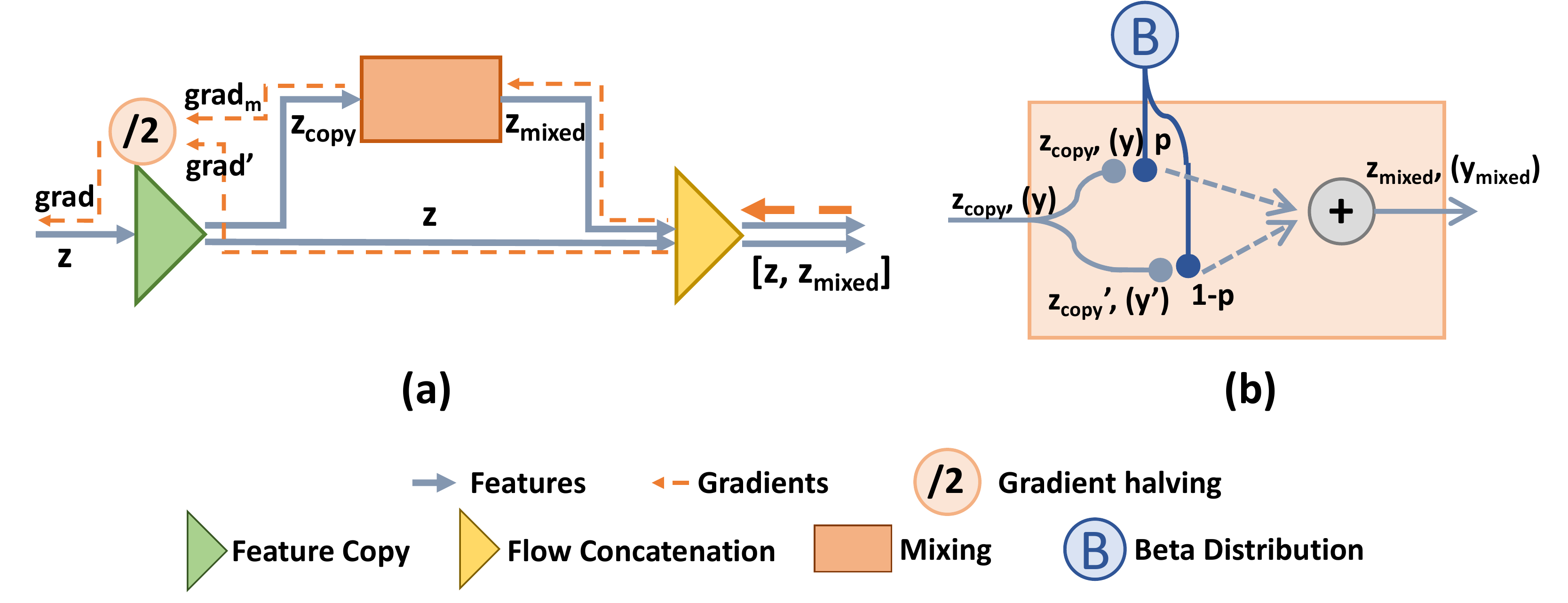}
    \caption{The left part shows the structure of the Mixing module. The right part gives the details of the mixing operation of Flow-Mixup. A copy of the feature maps is made and processed by the mixing operation, and then is batch-wise concatenated to the original ones.}
    \label{fig:mixingmodule}
\end{figure*}
\subsection{Preliminaries}
\indent Mixup~\cite{MIXUP} introduced a linear constraint to single-labeled classification and achieved good performance. Considering a deep learning classifier as a function $h(\cdot)$, the standard Mixup is defined as:
\begin{equation}\label{eq:mixing}
    h(px_p+qx_q) = py_p + qy_q
\end{equation}
where $x_p$ and $x_q$ are two input images while $y_p$ and $y_q$ are the corresponding labels, with $q=1-p$. Mixup regularization restricts the whole model (the function $h$) to be a linear function, as $h(px_p+qx_q) = py_p + qy_q = ph(x_p) + qh(x_q)$. Similarly, Manifold Mixup~\cite{manifoldmixup} applies the mixing operation as in Eq.~(\ref{eq:mixing}) to a hidden state, and restricts the subsequent parts of the model to learn a linear function. Note that the ``linear function'' and ``nonlinear function'' are different from the ``linear layer'' and ``nonlinear layer'' of the neural networks, as the former concepts are related to the learning objectives but the later concepts are about the model entities.
\subsection{An Overview of Flow-Mixup}
\indent In this paper, we propose a new regularization approach, Flow-Mixup, for multi-labeled medical image classification. Consider a deep learning classifier $h(x) = f(g(x))$, where $g$ is a nonlinear function and $f$ is a linear function. A training forward process with Flow-Mixup takes several steps: First, we select a hidden state $s$ to split the model into a nonlinear part and a linear part before training, as $s = g(x), y=f(s)$, and $y$ is the model output. Second, we process the data (e.g., the images) forward to the selected hidden state, and apply a new Mixing module to the features in the hidden state (our Mixing module is depicted below). After being processed by the Mixing module, the features continue the forward propagation until the output. With the Mixing module, Flow-Mixup restricts the front part of the model to learning a nonlinear function, and the rest of the model serves as a linear function. In dealing with multi-labeled medical images, the nonlinear function extracts abnormality-specific features, and the linear function (subsequent part) of the model projects the abnormality-specific features into the label spaces. The constraint to the nonlinear part is guaranteed, as the output of the nonlinear part is fed to the linear part which requires its input to lie in a linearly separable space. Different from Manifold Mixup, the special Mixing module introduces an extra flow dimension, thus simultaneously using several mixing modules in a model is allowed.
\subsection{Mixing Module}
\begin{figure*}[htbp]
    \centering
    \includegraphics[width=0.7\textwidth]{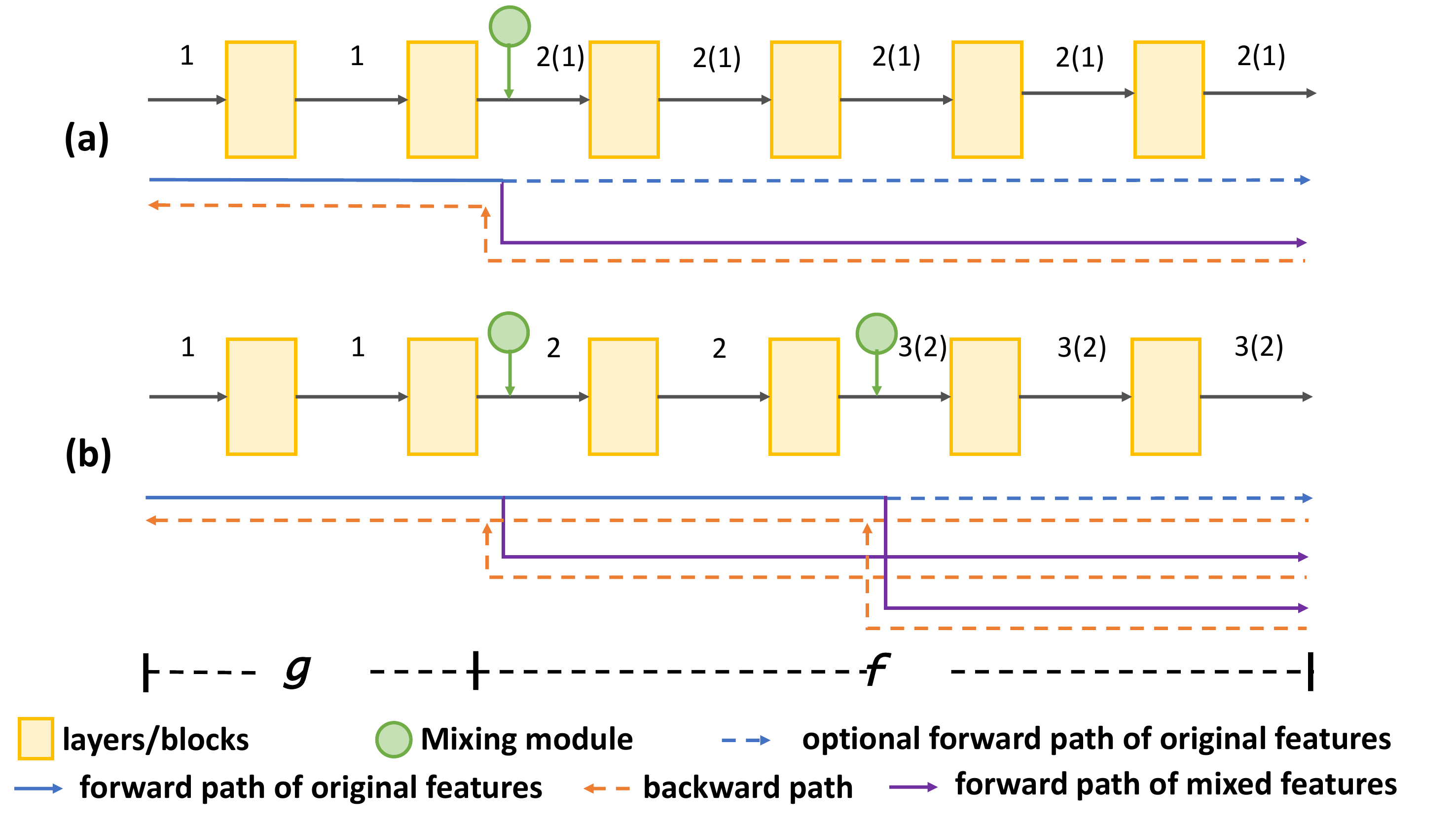}
    \caption{Illustrating two strategies to use Flow-Mixup. (a) The Mixing module is utilized only on a hidden state. (b) Mixing modules are applied to two hidden states. The numbers along the models indicate the flow dimension sizes, with and without the optional forward path of original features.}
    \label{fig:versions}
\end{figure*}

\indent Generally, the tensors of an image in deep learning models have 4 dimensions: batch dimension, channel dimension, width and height dimensions. Our proposed Flow-Mixup introduces a new dimension, called flow dimension. As shown in the left part of Fig.~\ref{fig:mixingmodule}, assume that the original feature $z$ has a flow dimension of size 1 before being processed by the Mixing module, and then the output of the Mixing module $[z, z_{\text{mixed}}]$ has a flow dimension of size 2. The flow size is increased by the feature concatenation operation. After the features are fed to the Mixing module, the first step is to make a copy of these features. Then, the feature copy is processed by a mixing operation and then concatenated into the original features along the flow dimension. The forward process in the Mixing module is defined as:
\begin{equation}\label{eq:forward}
z' = M(p, z) = [z, \text{Mixing}(p, z)] = [z, z_\text{mixed}]
\end{equation}
where the mixing operation $\text{Mixing}(\cdot,\cdot)$ transforms a feature copy $(z_\text{copy}, y)$ into two mini-copies ($(z, y),(z', y')$) and applies the standard Mixup to them by $(z_\text{mixed}, y_\text{mixed}) = (p z + (1-p)z', p y + (1-p)y')$, as illustrated in the right part of Fig.~\ref{fig:mixingmodule}. $(z, y)$ and $(z', y')$ are obtained by applying random index-shuffle to $(z_\text{copy}, y)$. $p$ is randomly sampled from a beta distribution $p_\mathcal{B}(\alpha, \alpha)$ and $\alpha$ is a hyper-parameter controlling the mixing degree~\cite{MIXUP}. $[\cdot, \cdot]$ indicates flow-wise concatenation, which results in the flow dimension size increase. Following Eq.~(\ref{eq:forward}), the feature $z$ is transformed into $z'$ with a double flow size. Since the flow size is doubled in the forward propagation, the Mixing module shall halve the gradients in the back-propagation in order to keep the magnitudes of the gradients. The back-propagation of the Mixing module is defined as:
\begin{equation}
    \text{grad} = (\text{grad}' + \text{grad}_m) / 2
\end{equation}
where $\text{grad}'$ indicates the gradients of the original features, and $\text{grad}_m$ represents the gradients of the mixed features (see Fig.~\ref{fig:mixingmodule}). In this way, the Mixing module can be applied to several hidden states simultaneously with the original features being preserved, as shown in Fig.~\ref{fig:versions}(b). Note that the regularization approach cannot entirely restrict the subsequent layers to be a linear function, and thus applying several Mixing modules is helpful in strengthening the linear constraints. In implementation, if a hidden state is the last one (see Fig.~\ref{fig:versions}(b)) or there is only one state (see Fig.~\ref{fig:versions}(a)) to apply the Mixing module, it is optional to compute the forward propagation of the original features to the output layer. If the original features do not go forward, the Mixing module degrades into the common Mixup operation, calculating $z' = z_\text{mixed}$ in the forward propagation and $\text{grad}=\text{grad}_m$ in the back-propagation.
\section{Analysis and Comparisons}\label{sec:analysis}
\indent This section discusses the feasibility and the characteristics of Flow-Mixup and its differences with the known regularization methods, Mixup~\cite{MIXUP} and Manifold Mixup~\cite{manifoldmixup}.
\subsection{Feasibility of Flow-Mixup}\label{sec:proof}
\noindent \textbf{Hypothesis 1:} {\it A learned sequential deep learning classifier for multi-labeled images can be reformulated as a composition of some linear functions and some nonlinear functions.}\\
\indent Since the feature correlations among abnormalities exist and the labels lie independently in the label space, a deep learning classifier needs to learn nonlinear functions in order to decouple the correlative features. Thus, it is reasonable to regard a learned sequential classifier as a composition of multiple nonlinear functions and linear functions. \\
\indent Based on Hypothesis 1, a learned sequential deep learning classifier $h(\cdot)$ can be mathematically decoupled by:
\begin{equation}\label{decouple_models}
    h(x) = f_1 \circ g_1 \circ f_2 \circ g_2 \circ \cdots \circ f_a \circ g_b(x)
\end{equation}
where the functions $f_i$ ($i \in \{1,2,\ldots,a\}$) and $g_j$ ($j \in \{1,2,\ldots,b\}$) belong to a linear function family $\mathcal{F}$ and a nonlinear function family $\mathcal{G}$, respectively. $f_i \circ g_j(x) = g_j(f_i(x))$. In practice, in Eq.~(\ref{decouple_models}), the order and the concrete expressions of $f_i$ and $g_j$ are obtained by learning from the data.\\
\textbf{Theorem 1.}\label{lemma1} {\it A learned sequential deep learning classifier $h(\cdot)$ for multi-labeled images can be reformulated as a composition of some linear functions and some nonlinear functions, in a sequence where the nonlinear functions $\tilde{g}_i$ appear first and then the linear functions $\tilde{f}_j$ follow, as:}\\
\begin{equation}
    h(x) = \tilde{g}_1 \circ \tilde{g}_2 \circ \cdots \circ \tilde{g}_c \circ \tilde{f}_1 \circ \tilde{f}_2 \circ \cdots \circ \tilde{f}_d (x) 
\end{equation}
\textbf{Proof:} A commutative law for linear and nonlinear functions can be proved, as follows. Assume $f \circ g(x) = y$ for a linear function $f$ and a nonlinear function $g$. Then $\exists \tilde{f} \in \mathcal{F}$ and $\exists \tilde{g} \in \mathcal{G}$ such that $f \circ g(x) = (f \circ g \circ \tilde{f}^{(-1)}) \circ \tilde{f}(x)$ = $\tilde{g} \circ \tilde{f}(x)$ = $y$ (because a linear function is invertible), with $\tilde{g} = f \circ g \circ \tilde{f}^{(-1)}$. Thus, $f \circ g(x) = \tilde{g} \circ \tilde{f}(x)$. By applying this commutative law repeatedly, it is easy to prove that a model $h(x)$ under Hypothesis 1 can be specified as:
\begin{equation}\label{eq:reformation_model}
    h(x) = \tilde{g}_1 \circ \tilde{g}_2 \circ \cdots \circ \tilde{g}_b \circ \tilde{f}_1 \circ \tilde{f}_2 \circ \cdots \circ \tilde{f}_a (x)
\end{equation}
where $d=a$ and $c=b$. The solution thus constructed (i.e., Eq.~(\ref{eq:reformation_model})) verifies the theorem, which suggests that any multi-labeled image classifier can find a solution under the constraint of Flow-Mixup if the equation of the original classifier has a solution. \hfill $\square$
\subsection{Comparisons with Mixup}\label{sec:mixup}
\begin{figure}[tb]
\centering
\includegraphics[width=0.8\textwidth]{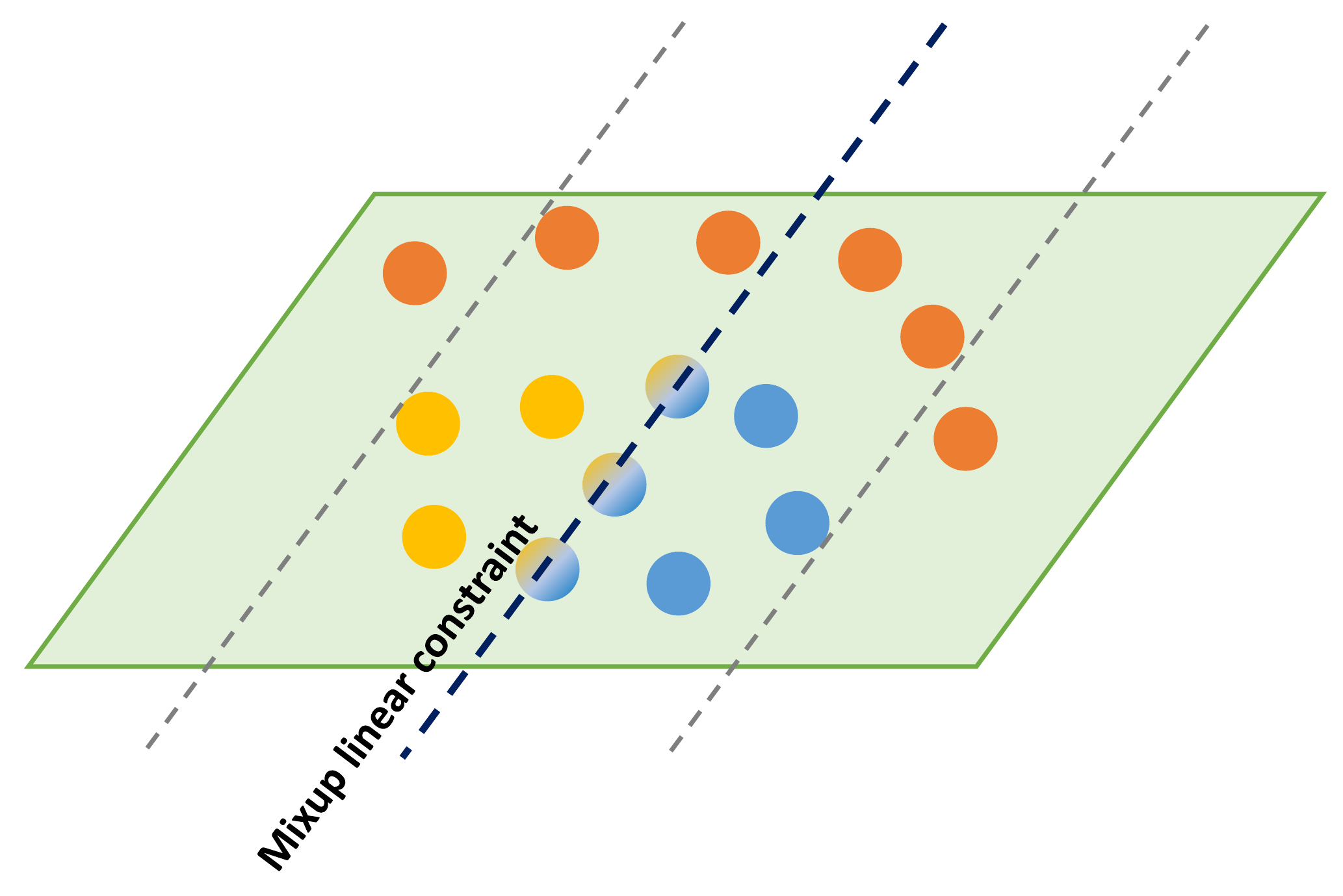}
\caption{Illustrating ``correlation conflicts''. Assume that the yellow balls and blue balls indicate samples with two different labels, while the orange balls are for samples with both of the two labels. One can see the Mixup linear constraint conflicting with the original correlation.}\label{fig:comparison_mixup}
\end{figure}

\indent As discussed in the previous work~\cite{graphx}, the features of abnormalities can be correlative, which may not be linearly separable. In other words, the inherent correlation of abnormalities might be in conflict with the linear constraint of Mixup. Thus, training a multi-labeled image classifier with the Mixup regularization may result in performance decrease. As the situation illustrated in Fig.~\ref{fig:comparison_mixup}, such ``correlation conflicts'' happen as the boundary line of two classes cannot deal with the data belonging to both of these two classes, after mapping the data manifold to a low dimensional space satisfying the Mixup linear constraint. In contrast, with our Flow-Mixup, the correlative features of abnormalities can be decoupled into abnormality-specific features by the nonlinear functions first, and such features lie in a linearly separable space.
\subsection{Comparisons with Manifold Mixup}\label{sec:manifold}
\indent Manifold Mixup~\cite{manifoldmixup} allows applying a mixing operation to several hidden states in the training process. However, this mixing operation cannot be performed simultaneously. Manifold Mixup randomly selects one of these hidden states to apply the mixing operation in every training iteration, and consequently suffers two drawbacks. (1) Updating parameters in every iteration affects the final parameters. Therefore, it is hard to know exactly what degree of data mixing is applied to a hidden state, as the mixing operation is used with a probability. Thus, it is difficult to determine the hyper-parameters for the mixing operation. (2) Since the training condition to a hidden state (whether to use a mixing operation) is changeable, the training process is unstable and suffers a ``distribution shift'' phenomenon. ``Distribution shift'' means that the objective feature distribution is changed. Ideally, using a mixing operation on a hidden state restricts the features to lie in a linearly separable space. However, Manifold Mixup keeps changing the constraint to the hidden states, which leads to an unstable training process and decreases the performance.\\
\begin{figure}[tb]
\centering
\includegraphics[width=0.8\textwidth]{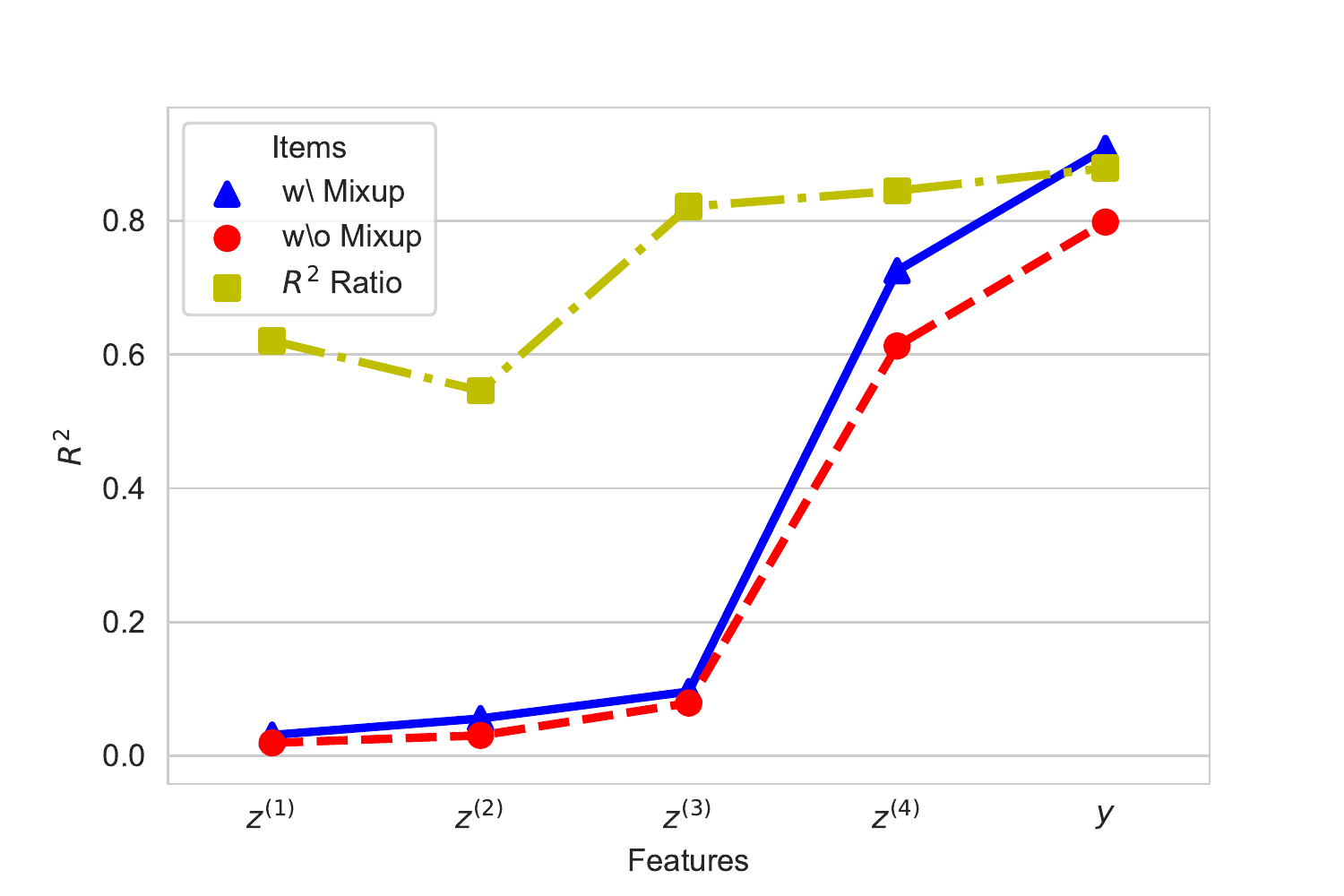}
\caption{A visualization of the $\text{R}^2$ values of different hidden states with or without Mixup on the training set of CIFAR-10. The $\text{R}^2 \text{ Ratio}=\text{R}_1^2/\text{R}_2^2$, where $\text{R}_1^2$ is $\text{R}^2$ values without Mixup and $\text{R}_2^2$ is $\text{R}^2$ values with Mixup. One can see that the feature distributions with and without Mixup are quite different.}\label{fig:comparison_manifold}
\end{figure}
\indent To observe the occurrence of the ``distribution shift'' phenomenon in model training, we compare the feature distributions on the training set of CIFAR-10, as shown in Fig.~\ref{fig:comparison_manifold}. We train the PreAct-ResNet-32 model~\cite{he2016identity} on the training set of CIFAR-10 with Mixup (applied to the data input and the output of every residual block with $p_\mathcal{B}(1.0,1.0)$) and without Mixup. Then we collect the output of every residual block and the model output. To avoid the influence of the classification results, we utilize the k-means clustering algorithm (partitioning into $k=10$ classes) on the collected features of every block output and model output. Then we calculate the average value of $\text{R}^2$ (similar to $\text{R}^2$ in the analysis of variance) to observe the feature distributions. $\text{R}^{2}=1 -\text{SSI}/\text{SST}$, where $\text{SSI}$ is the sum of squares for intra-cluster and $\text{SST}$ is the total sum of squares. $\text{R}^{2}$ presents the percentage of the total variance coming from the inter-cluster variance. The higher $\text{R}^{2}$ is, the more clear the boundaries of the clusters are. SSI and SST are defined by:
\begin{equation}
\begin{cases}
SST^{(i)} = \frac{1}{V^{(i)}} \sum^{N}_{j=0}\sum^{V^{(i)}}_{v=1} (z_{j,v}^{(i)} - \Bar{\Bar{z}}_{v}^{(i)})^2\\
SSI^{(i)} = \frac{1}{V^{(i)}}\sum^{C}_{c=0} \sum^{N_c}_{j=1} \sum^{V^{(i)}}_{v=1} (z_{c,j,v}^{(i)} - \Bar{z}_{c,v}^{(i)})^2
\end{cases}
\end{equation}
where $C$ indicates the number of clusters, $N$ is the number of images, and $N_c$ is the number of the images belonging to the $c$-th cluster. $z_j^{(i)}$ is the features of the $j$-th images in the $i$-th hidden state. $V^{(i)}$ denotes the feature size of one data in the $i$-th hidden state, i.e., $V=D \times H \times W$, where $D,H$, and $W$ are the channel, height, and width dimension sizes. $\Bar{\Bar{z}}^{(i)}$ and $\Bar{z}_c^{(i)}$ denote the data-wise average features in the $i$-th hidden state and the data-wise average features of the $c$-th cluster in the $i$-th hidden state, respectively. As shown in Fig.~\ref{fig:comparison_manifold}, one can see that $\text{R}^{2}$ of the features learned with Mixup is evidently higher than without any mixing operations. Thus, the ``distribution shift'' phenomenon happens when using Manifold Mixup, as the objective feature distributions are very different with and without mixing operations.
\section{Experiments}
\begin{table*}[tb]
\begin{tabular}{l|cccccccccccc}
\hline\hline
\multirow{2}{*}{Regularization}       & \multicolumn{3}{c}{$0\% \times$} & \multicolumn{3}{c}{$10\% \times$} & \multicolumn{3}{c}{$30\% \times$} & \multicolumn{3}{c}{$50\% \times$} \\ \cline{2-13}
                                        & Best       & Last   & Diff.    & Best    & Last   & Diff.   & Best & Last & Diff.      & Best        & Last  & Diff.    \\ \hline
ERM   & 74.2 & 69.4 & 4.8 & 73.1 & 69.9 & 3.2 & 72.8 & 69.3 & 3.5 & 72.2 & 67.3 & 4.9\\
Mixup ($\alpha$ = 1.0)  & 77.0 & 76.2 & 0.8 & 76.8  & \underline{76.4} & 0.4 & 76.4 & \textbf{75.9} & 0.5 & 75.7 & 74.6 & 1.1\\
Mixup ($\alpha$ = 3.0) & 76.7  & 75.8 & 0.9 & 76.3  & 75.6 & 0.7 & 76.3 & 75.6 & 0.7 & \underline{76.1}  & \textbf{75.5} & 0.6\\
Manifold Mixup ($\alpha$ = 3.0)  & \underline{77.3}  &  75.5 & 1.8 & 76.6  & 75.3 & 1.3 & \underline{76.6}   & 74.8 & 1.8  & 75.8  & 74.3  & 1.5 \\\hline
Flow-Mixup ($\alpha$=3.0, Op=False)  & \textbf{77.8}  & \textbf{76.9} & 0.9 & \textbf{77.1} & 76.3 & 0.8 &76.5 & 75.3 & 1.2 & \underline{76.1} & 75.1 & 1.0\\
Flow-Mixup ($\alpha$=3.0, Op=True) & 76.9 & \underline{76.5} & 0.4 & \underline{76.9}  & \textbf{76.7} & 0.2 & \textbf{77.0}  & \underline{75.7}& 1.3  & \textbf{76.3} & \underline{75.4} & 0.9\\ \hline \hline
\end{tabular}
\caption{Comparison of regularization methods using DenseNet-121 and in AUC ($\%$) on the ChestX-ray14 test set with label corruption. Here, $\alpha$ is the hyper-parameter of the beta distributions in the Mixing operation, and ``Op'' indicates whether the original features going forward (see Fig.~\ref{fig:versions}). The best and the second best results are marked as \textbf{bold} and \underline{underlined}, respectively.}\label{exp:AUC}
\end{table*}
\begin{table*}[tb]
\begin{tabular}{c|l|cccc}
\hline\hline
Dataset & Regularization  & $0\% \times$ & $10\% \times$ & $40\% \times$ & $70\% \times$ \\ \hline
\multirow{6}{*}{ECG-12} & ERM & 0.6617 & 0.6531 & 0.6238 & 0.5590\\
& Mixup ($\alpha$ = 1.0) & 0.6773 & 0.6581 & 0.6337 & 0.5774\\
& Mixup ($\alpha$ = 3.0) & 0.6575 & 0.6225 & 0.6195 & 0.5894 \\
& Manifold Mixup ($\alpha$ = 3.0) & 0.6436 & 0.6389 & \underline{0.6378} & 0.5822\\ \cline{2-6}
& Flow-Mixup ($\alpha$=3.0, Op=False) & \textbf{0.6994} & \textbf{0.6800} & 0.6347 & \textbf{0.5996} \\
& Flow-Mixup ($\alpha$=3.0, Op=True) & \underline{0.6846} & \underline{0.6784} & \textbf{0.6495} & \underline{0.5963} \\ \hline\hline
\multirow{6}{*}{ECG-55} & ERM & 0.5535 & 0.5319 & 0.4501 & 0.3280\\
& Mixup ($\alpha$ = 1.0) & 0.5543 & 0.5119	& 0.4992 & 0.3474\\
& Mixup ($\alpha$ = 3.0) & 0.5509 & 0.5245 & 0.4953 & 0.4563 \\
& Manifold Mixup ($\alpha$ = 3.0) & 0.5512 & 0.5390 & 0.4966 & \textbf{0.4655}\\
\cline{2-6}
& Flow-Mixup ($\alpha$=3.0, Op=False) & \underline{0.5551} & \underline{0.5416} & \textbf{0.5113} & 0.4426\\
& Flow-Mixup ($\alpha$=3.0, Op=True) & \textbf{0.5769} & \textbf{0.5454} & \underline{0.5072} & \underline{0.4570} \\ \hline \hline
\end{tabular}
\caption{comparison of regularization methods using 1D-ResNet-34 and in Macro-F1 ($\%$) on the ECG test sets with label corruption. Here, $\alpha$ is the hyper-parameter of the beta distributions in the Mixing operation, and ``Op'' indicates whether the original features going forward (see Fig.~\ref{fig:versions}). The best and the second best results are marked as \textbf{bold} and \underline{underlined}, respectively.}\label{exp:macrof1}
\end{table*}
\subsection{Datasets}
\begin{figure}[tb]
    \centering
    \includegraphics[width=1.0\textwidth]{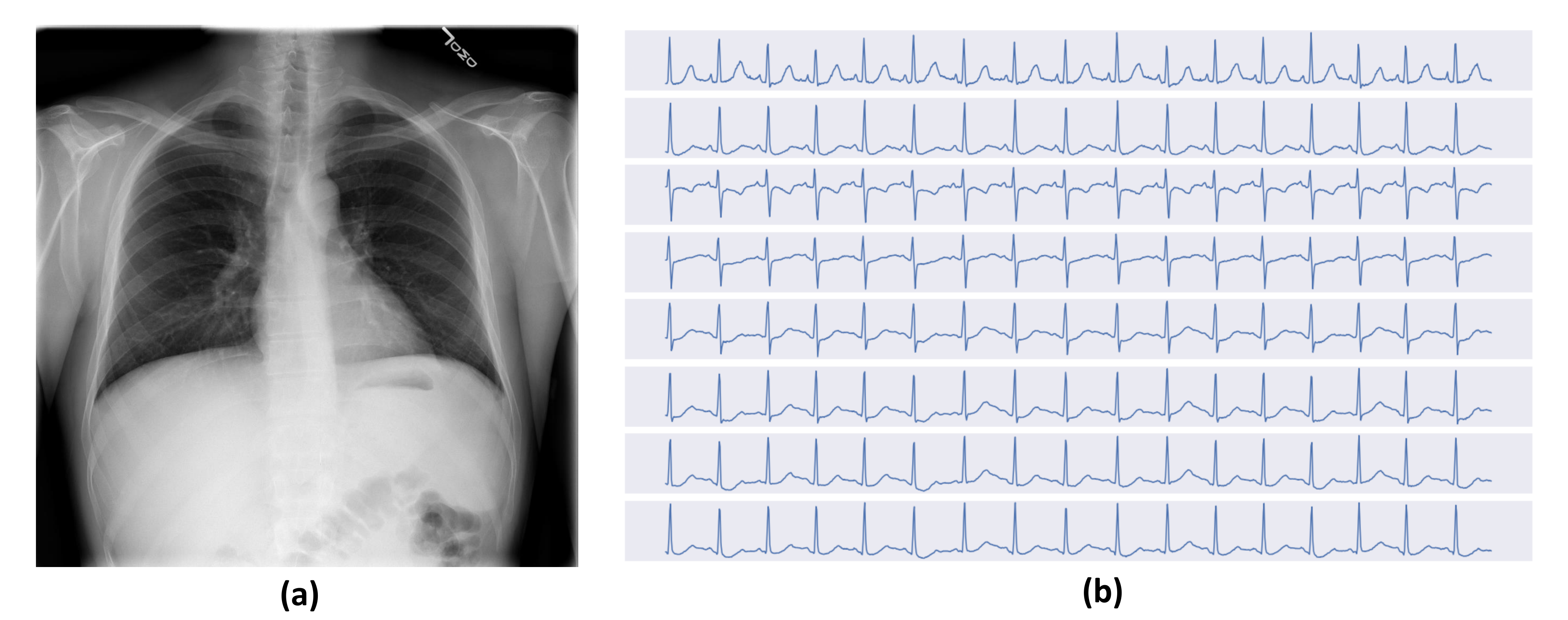}
    \caption{(a) A chest X-ray image from the ChestX-ray14 dataset. (b) An 8-lead electrocardiogram record from the ECG datasets.}
    \label{fig:example}
\end{figure}

\indent To evaluate our Flow-Mixup approach for multi-labeled medical image classification tasks, we conduct experiments on the ChestX-ray14 dataset~\cite{Wang2017CVPR} and two ECG record datasets of the Alibaba Tianchi Cloud Competition\footnote{\url{https://tianchi.aliyun.com/competition/entrance/231754/introduction}}. These datasets are for multi-labeled medical image classification. The ChestX-ray14 dataset~\cite{Wang2017CVPR} consists of 112,120 CXR images of size $1,024 \times 1,024$ each. The corresponding labels cover 14 abnormalities extracted from radiology reports by natural language processing (NLP), and some of the CXR images are assigned with more than one label. As estimated by data collectors~\cite{Wang2017CVPR}, there is $\sim$10\% label corruption. For the ECG classification, we use the preliminary competition ECG dataset (the ECG-55 dataset) containing 55 arrhythmia categories, and a selected ECG dataset (the ECG-12 dataset) containing 12 most common arrhythmia categories in which the ECG records are selected from the preliminary dataset and the final competition dataset. The ECG-55 dataset consists of 31,779 8-lead ECG records and ECG-12 has 34,664 8-lead ECG records. The ECGs are 10 second records and were recorded at a frequency of 500 Hertz. An ECG record can be treated as a special one-dimensional image and with 1-D convolutions~\cite{kachuee2018ecg,shen2019ambulatory}. Example samples of the ChestX-ray14 and ECG datasets are shown in Fig.~\ref{fig:example}. In experiments, for the ChestX-ray14 dataset, we follow the official split, and for the ECG datasets, we randomly split a dataset into training, validation, and test parts by 7:1:2 since the official test set is not available.
\subsection{Experimental Setups}
\indent We use DenseNet-121~\cite{huang2017densely} as the CXR classifier baseline and ResNet-34~\cite{he2016identity} as the ECG classifier baseline. Two convolutional layers are added ahead of the DenseNet-121 network as was done similarly in~\cite{DNetLoc}, both with a kernel size of 3 and a stride of 2. For CXR image classification, we follow the weighted binary cross-entropy loss function~\cite{DNetLoc,Wang2017CVPR}, weighting the loss term for an abnormality with its inverse proportion. In ResNet-34, 1-D convolution kernels of size 3 are used to replace the $3 \times 3$ convolution kernels. During training, we set the batch size as 32, and employ the Adam optimizer~\cite{Adam} with $\beta_1=0.9$ and $\beta_2=0.999$. The learning rate is initialized as $10^{-4}$ and is reduced by $10 \times$ when the valid loss reaches a plateaus. We run 50 epochs for CXR image classification and 200 epochs for ECG classification. To validate Flow-Mixup and compare to the known regularization methods on multi-labeled classification with label corruption, we replace the labels with corrupted labels in probability. The label corruption rates for the ECG reports are $0\%\times$, $10\%\times$, $40\%\times$, and $70\%\times$, while for CXR images the label corruption rates are $0\%\times$, $10\%\times$, $30\%\times$, and $50\%\times$ as the original labels are already corrupted. The mixing operation is applied to the input of the third and fifth ResBlock and Denseblock, for both Manifold Mixup and Flow-Mixup. We report the average AUC (Area under the ROC curve) over the 14 kinds of abnormalities on the CXR test set, and report Macro-F1 (macro-averaging on F1 scores) on the two ECG test sets.
\begin{figure*}[tb]
    \centering
    \includegraphics[width=1.0\textwidth]{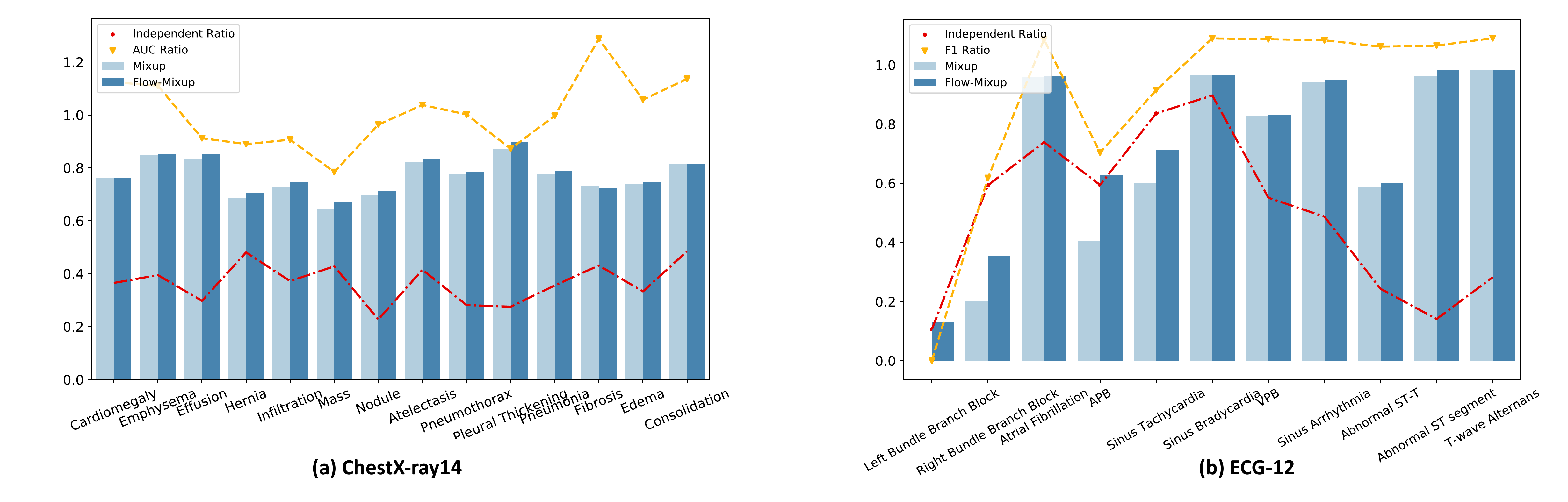}
    \caption{Illustrating the AUC and F1 score for every abnormality with Mixup ($\alpha$ = 3.0) and Flow-Mixup ($\alpha$ = 3.0, Op = False). In Subfig. (b), we show just 11 abnormalitiy categories except for a normal category. Note that Mixup achieves F1 scores = 0.0 in the left bundle branch block abnormality.}
    \label{fig:AUC_details}
\end{figure*}
\subsection{Experimental Results}
\subsubsection{Performance Comparison}
The experimental results on the ChestX-ray14 dataset are reported in Table~\ref{exp:AUC} and the results on the ECG-55 and ECG-12 datasets are in Table~\ref{exp:macrof1}. We compare the indicators of our proposed Flow-Mixup with those of several known state-of-the-art regularization methods, including the Empirical Risk Minimization (ERM) principle~\cite{ERM}, Mixup~\cite{MIXUP}, and Manifold Mixup~\cite{manifoldmixup}. We report the best and the last performances in CXR classification; we report only the best performances in ECG classification, since the last performances are very close to the best performances on the ECG datasets. One can see that Flow-Mixup outperforms the other regularization methods in dealing with various degrees of label corruption. Flow-Mixup's performance over the other regularization methods validates the capability of Flow-Mixup. Flow-Mixup attains better performances than Mixup, which might result from the abnormality-specific features extracted by the nonlinear part (see Sec.~\ref{sec:mixup}). In ECG classification, Flow-Mixup outperforms Mixup and Manifold Mixup, and a similar conclusion can be derived. Further, one can see from Fig.~\ref{fig:AUC_details} that Flow-Mixup outperforms Mixup in most classes in F1 scores and AUCs.
\subsubsection{Correlation Conflict Reduction}
To further evaluate Flow-Mixup's ability to reduce correlation conflicts, we compare the F1 score and AUC of every abnormality between Mixup and Flow-Mixup on the ChestX-ray14 test set and ECG-12 test set, respectively. The histograms of these F1 scores and AUCs are shown in Fig.~\ref{fig:AUC_details}, both of them with $\sim$10\% label corruption. For easy comparison, we set two new indicators for every class: ``Performance Ratio'' = $(\text{Performance}_{\text{Mixup}} \ / \ \text{Performance}_{\text{Flow-Mixup}})^{r}$ ($r$ is an exponent; $r=10.0$ for CXR images and $r=1.0$ for ECG records), and ``Independent Ratio'' = $n_c/m_c$, where $n_c$ is the number of images with only the class $c$, and $m_c$ is the number of all the images with the class $c$ including multi-labeled images. The ``Performance Ratio'' (which is the ``AUC Ratio'' for CXRs and ``F1 Ratio'' for ECGs in Fig.~\ref{fig:AUC_details}) indicates the relative performance in every abnormality of Mixup and Flow-Mixup, while the ``Independent Ratio'' suggests in what degree a class is independent in a dataset. The performances are normalized before computing the ``Performance Ratio''. Thus, one can see whether the relative performances are related to the class independence by comparing the coincides of the ``Performance Ratio'' and the ``Independent Ratio''. In Fig.~\ref{fig:AUC_details}, the ``Performance Ratio'' curves coincide with the ``Independent Ratio'' curves (with Spearman correlation coefficients $\approx 0.3$), indicating that Flow-Mixup can obtain better performance in a relatively dependent class. Hence, we believe Flow-Mixup is able to reduce the correlation conflicts.
\subsubsection{Distribution Shift Reduction}
\indent To evaluate Flow-Mixup's ability to reduce the ``Distribution Shift'' phenomenon, we compute the differences (Diff.s) between the Best AUCs and the Last AUCs on the ChestX-ray14 test set, shown in Table~\ref{exp:var_compare}. The Diff.s on the ECG test sets are not reported since the best and the last Marco F1 scores are very close. Further, we compute the variances of the normalized performance indicators (AUCs for CXR images and Macro F1 scores for ECG records) of some epochs on the test sets, as:
\begin{equation}
    \text{Var}(I) = \frac{\sum^n_{e=1}(I_e - \bar{I})^2}{n}
\end{equation}
where $I$ is the normalized performance on the test set, $\bar{I}$ is the average performance, and $e$ is the index of epochs. The normalization method is the min-max normalization. The variances are computed for the early 20 epochs on the ChestX-ray14 dataset ($n=20$) and for the early 100 epochs on the two ECG datasets ($n=100$), as the indicators just fluctuate slightly in the rest epochs. As shown in Table~\ref{exp:var_compare}, Flow-Mixup has lower variances than Manifold Mixup, which suggests that Flow-Mixup is more stable. Comparing the Diff.s and variances, it is obvious that training models with Manifold Mixup is not as stable as with Flow-Mixup, which might be due to the instability caused by ``distribution shift'', as discussed in Sec.~\ref{sec:manifold}.
\subsubsection{Hyperparameter $\alpha$}
As the suggestions in~\cite{MIXUP,manifoldmixup}, setting the mixing degree $\alpha>1.0$ is suggested in dealing with the corrupted labels. In our tasks, we find that the models perform well with $\alpha>1.0$. Flow-Mixup seems to be insensitive to $\alpha$, and the results fluctuate within $\sim 0.1$ on the ECG datasets and within $\sim 1.0$ on the ChestX-ray dataset with different $\alpha \in [1.0, 3.0]$.
\begin{table*}[tb]
\caption{Variances of the normalized performance indicators for Manifold Mixup ($\alpha$ = 3.0) and Flow-Mixup ($\alpha$ = 3.0, Op = False) over $n$ epochs on the test sets: AUC variances for the ChestX-ray14 test set and Macro-F1 variances for the ECG-55 and ECG-12 test sets. The lower variances are marked as \textbf{bold}.}\label{exp:var_compare}
\begin{tabular}{l|cccc|cccc|cccc}
\hline \hline
      & \multicolumn{4}{c|}{ChestX-ray14 ($n=20$)} & \multicolumn{4}{c|}{ECG-12 ($n=100$)} & \multicolumn{4}{c}{ECG-55 ($n=100$)} \\ \hline
      Regularization & $0\% \times$   & $10\% \times$ & $30\% \times$  & $50\% \times$  & $0\% \times$ & $10\% \times$ & $40\% \times$  & $70\% \times$  & $0\% \times$  & $10\% \times$  & $40\% \times$  & $70\% \times$  \\ \hline
Manifold Mixup &   \textbf{0.0636}   &    0.0891   & 0.1208  &  0.0909  &  0.0350 & 0.0336 &  0.0233  &  0.0316  & \textbf{0.0376} &   0.0538  &  0.0422  & 0.0276 \\ \hline
Flow-Mixup     &   0.0745    &    \textbf{0.0865}    &  \textbf{0.0732}   & \textbf{0.0714} & \textbf{0.0310} &  \textbf{0.0280} &   \textbf{0.0196}   & \textbf{0.0237} & 0.0408 &  \textbf{0.0530}  &  \textbf{0.0374} &  \textbf{0.0270} \\ \hline \hline
\end{tabular}
\end{table*}
\section{Conclusions}
\indent In this paper, we proposed a new regularization approach, Flow-Mixup, for multi-labeled medical image classification with corrupted labels. Guided by Flow-Mixup, a deep learning classifier extracts abnormality-specific features and then maps such features into the label space. Experiments verified that Flow-Mixup can handle datasets containing corrupted labels, and thus makes it possible to apply automatic annotation. Besides, we compared Flow-Mixup with the common Mixup and Manifold Mixup methods, highlighted the characteristics of Flow-Mixup, and discussed the ``correlation conflicts'' phenomenon and ``distribution shift'' phenomenon occurred with using Mixup or Manifold Mixup.
\section{Acknowledgements}
This research was partially supported by the National Research and Development Program of China under grant No. 2019YFB1404802, No. 2019YFC0118802, and No. 2018AAA0102102, the National Natural Science Foundation of China under grant No. 61672453, the Zhejiang University Education Foundation under grants No. K18-511120-004, No. K17-511120-017, and No. K17-518051-02, the Zhejiang public welfare technology research project under grant No. LGF20F020013, and the Key Laboratory of Medical Neurobiology of Zhejiang Province. D. Z. Chen’s research was supported in part by NSF Grant CCF-1617735.
\bibliographystyle{plain}
\bibliography{references}
\end{document}